\documentclass[10pt,twocolumn,letterpaper]{article}

\usepackage{iccv}
\usepackage{times}
\usepackage{epsfig}
\usepackage{graphicx}
\usepackage{amsmath}
\usepackage{amssymb}
\usepackage{todonotes}
\usepackage{algorithm}
\usepackage{algorithmic}
\usepackage{bm}
\usepackage{threeparttable}
\usepackage{multirow}
\usepackage{booktabs}

\newcommand{\beginsupplement}{%
        \setcounter{table}{0}
        \renewcommand{\thetable}{S\arabic{table}}%
        \setcounter{figure}{0}
        \renewcommand{\thefigure}{S\arabic{figure}}%
     }

\usepackage[breaklinks=true,bookmarks=false]{hyperref}

\iccvfinalcopy 

\def\iccvPaperID{****} 

\ificcvfinal\fi

\begin{document}

\title{Unleashing the Potential of Spiking Neural Networks with Dynamic Confidence}

\author{Chen Li\thanks{These authors contributed equally to this work.} \\
The University of Manchester\\
Manchester, United Kingdom\\
{\tt\small chen.li@manchester.ac.uk}
\and
Edward G Jones\footnotemark[1] \\
The University of Manchester\\
Manchester, United Kingdom\\
{\tt\small edward.jones-3@manchester.ac.uk}
\and
Steve Furber\\
The University of Manchester\\
Manchester, United Kingdom\\
{\tt\small steve.furber@manchester.ac.uk}
}

\maketitle
\ificcvfinal\thispagestyle{empty}\fi

\begin{abstract}

This paper presents a new methodology to alleviate the fundamental trade-off between accuracy and latency in spiking neural networks (SNNs). The approach involves decoding confidence information over time from the SNN outputs and using it to develop a decision-making agent that can dynamically determine when to terminate each inference. 

The proposed method, Dynamic Confidence, provides several significant benefits to SNNs. 1. It can effectively optimize latency dynamically at runtime, setting it apart from many existing low-latency SNN algorithms. Our experiments on CIFAR-10 and ImageNet datasets have demonstrated an average 40\% speedup across eight different settings after applying Dynamic Confidence. 2. The decision-making agent in Dynamic Confidence is straightforward to construct and highly robust in parameter space, making it extremely easy to implement. 3. The proposed method enables visualizing the potential of any given SNN, which sets a target for current SNNs to approach. For instance, if an SNN can terminate at the most appropriate time point for each input sample, a ResNet-50 SNN can achieve an accuracy as high as 82.47\% on ImageNet within just 4.71 time steps on average. Unlocking the potential of SNNs needs a highly-reliable decision-making agent to be constructed and fed with a high-quality estimation of ground truth. In this regard, Dynamic Confidence represents a meaningful step toward realizing the potential of SNNs. Code is available$\footnote{\url{https://github.com/chenlicodebank/Dynamic-Confidence-in-Spiking-Neural-Networks}}$.
\end{abstract}

\section{Introduction}
\label{sec:intro}

Deep artificial neural networks (ANNs) have achieved remarkable success in computer vision tasks~\cite{krizhevsky2012imagenet, redmon2016you, long2015fully}. These improvements have been accompanied by ever-growing model complexity and neural network depth. Though the classification accuracy has improved dramatically, the latency and energy cost have also increased, which poses a challenge for real-world AI applications on edge devices, such as mobile phones, smartwatches, and IoT hardware.

\begin{figure}
\centering
\includegraphics[width=0.465\textwidth]{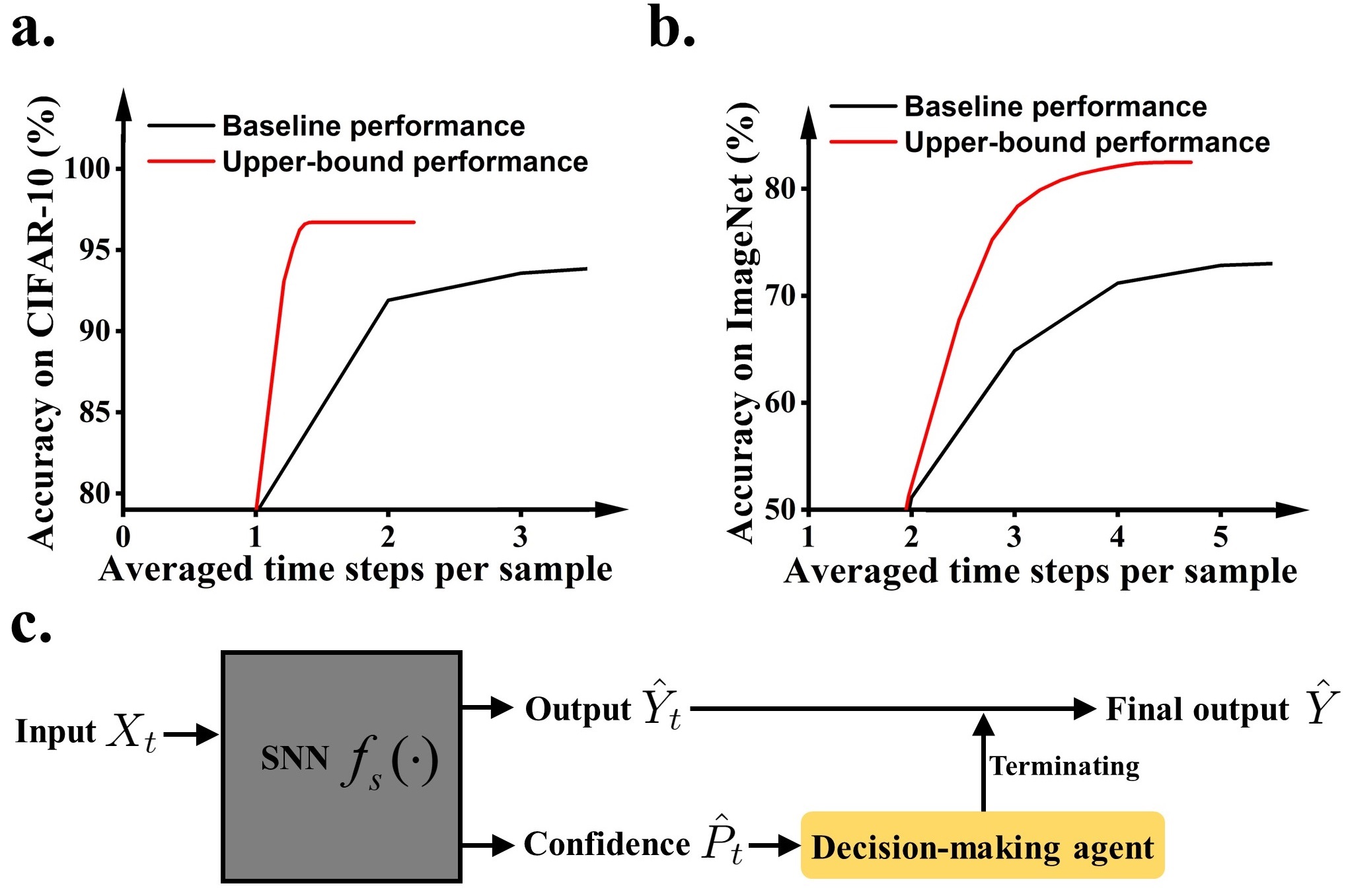} 
\caption{The upper-bound performance of SNNs when fully utilizing dynamic strategies at runtime, as shown by the red curves in \textbf{a.} ResNet18 on CIFAR-10 and \textbf{b.} ResNet-50 on ImageNet. The black curves represent baseline SNN performance without dynamic strategies. Additional figures with other settings can be found in the Supplementary Material.  \textbf{c.} The diagram of the proposed Dynamic Confidence, which can be implemented on-the-fly.}
\label{fig1}
\end{figure}


Deep spiking neural networks (SNNs) are promising to offer power and latency advantages over counterpart deep artificial neural networks during inference. However, when their power (e.g. averaged spike counts per inference) or latency (e.g. averaged inference time steps per inference) are strictly constrained, SNNs suffer huge accuracy degradation. In recent years, much scholarly attention has been paid to the conflict between accuracy and latency in SNNs, growing the field of low-latency SNNs (also known as fast SNNs). The primary goal of fast SNN research is achieving equivalent or slightly-lower accuracy than the baseline ANN accuracy with as few inference time steps as possible. With the reduction in the number of  inference time steps, the spike counts required for each inference can potentially be reduced as well, if the firing rates can be kept the same when latency decreases. In hardware that can take scale power with spike processing operations, this research on low-latency SNNs can contribute to reducing the power requirements of SNNs.

In this paper, we detail a runtime optimization method called Dynamic Confidence and show that it can effectively reduce the inference latency in low-latency SNNs on CIFAR-10 and ImageNet, without the need to sacrifice accuracy or to increase firing rates. Our paper has several new contributions:

\begin{itemize}
    \item  We propose a method (Dynamic Confidence) to  decode a temporal series of confidence values from the inference results of an SNN, and utilize this confidence information to dynamically guide inference. It is the first study to formulate confidence in SNNs and introduce the formal concept of dynamic strategies to SNNs.
    
    \item Dynamic Confidence can further reduce the inference latency by 40\% on average on the state-of-the-art low-latency SNNs (QFFS\cite{li2022quantization} and QCFS\cite{bu2021optimal}) on CIFAR-10 and ImageNet. By reducing the inference latency and maintaining the same firing rates, the power consumption of SNN inference is reduced consequently.

    \item We provide a method to calculate the upper-bound performance that an SNN can achieve when applying dynamic strategies at runtime. Our findings shed light on the vast untapped potential within current SNNs, providing valuable insights to answer the fundamental question of why to use SNNs instead of ANNs.

    \item Dynamic Confidence is a lightweight, parameter-insensitive, on-the-fly method. These features make Dynamic Confidence extremely easy to be applied on a variety of low-latency SNN algorithms whenever further latency and power reductions are desirable. 
\end{itemize}

\section{Motivation}

\begin{itemize}
    \item A main research method adopted in current SNN studies is improving SNNs by leveraging knowledge from ANNs. Some successful examples of this include, but are not limited to, straight-through estimators~\cite{neftci2017event}, quantization~\cite{li2022quantization}, Backpropagation through time (BPTT)~\cite{bellec2020solution}, network architectures~\cite{li2022converting}, neural architecture search ~\cite{kim2022neural}, and neural network simulation tools ~\cite{wu2019direct}. The primary concern during this process is how to make ANN knowledge applicable to SNNs, which usually requires additional optimization of spike dynamics in SNNs to ensure smooth knowledge transfer. The study presented in this paper also follows this research method. Specifically, this study is intended to explore whether confidence, a key concept in Bayesian neural networks and interpretable AI can be applied to SNNs to help reduce inference latency as well as inference power. We discovered that the temporal dimension of SNNs provides richer information in confidence compared to ANNs. This paper  exemplifies a simple way to leverage the information in SNN confidence to further boost the performance of SNNs.

    \item Dynamic strategies have received considerable critical attention in ANNs due to their effectiveness in bringing considerable speedup and computational savings without degrading prediction accuracy ~\cite{wu2018blockdrop}.
    In contrast, there is not any research studying the application of dynamic strategies to SNNs and the evaluation of possible speed and power gains on nontrivial datasets such as CIFAR-10 and ImageNet. This paper seeks to address an existing gap by sharing the preliminary findings of implementing a dynamic strategy in SNNs. Different from
    other dynamic strategy research in ANNs, we found in this paper that
    spikes make it convenient to apply a dynamic strategy explicitly 
    in the dimension of precision. This task has been notably challenging in existing ANN dynamic strategy literature. 
    \item Information is in the central position in SNN research. For example, a primary concern in directly-trained SNNs is whether better spatial-temporal information representation and computing can be achieved by surrogate gradients. This paper also emphasizes the significance of the information, but from a different perspective. Instead of seeking more spatial-temporal information by replacing ANN-to-SNN conversion with spatial-temporal training by surrogate gradients, we stick to using ANN-to-SNN conversion but try to seek if any information in the SNN output is missed by other research and has not been fully exploited. Our results show that confidence in SNN outputs contains valuable information and can be leveraged to improve SNN performance on latency and power.
\end{itemize}

\section{Related Work}

\noindent \textbf{Spiking Neural Networks (SNNs)}. Spiking neural networks (SNNs) are biologically-inspired models that have time-evolving states, unlike artificial neural networks (ANNs)~\cite{maass1997networks, li2021towards, ding2022snn, ding2021accelerating}. SNNs use spikes, binary pulses that model action potentials in biology, to transmit information between neurons. Because data processing in SNNs need only take place when spikes are received, SNNs can be implemented in an event-based manner and can benefit from improved power efficiency over ANNs, especially when sparsity can be exploited in hardware~\cite{kimSpikingYOLOSpikingNeural2020, dampfhofferAreSNNsReally2022, juFPGAImplementationDeep2020}. 

The potential for increased efficiency and biological plausibility that SNNs promise has led to ANN-to-SNN conversion being an active area of research.  Converting ANN activations to spike rates in SNNs has proven possible~\cite{diehl2015fast} with marginal loss in accuracy when data-based normalisation techniques are applied~\cite{rueckauer2017conversion}. The significant downsides to these kinds of rate-coding approaches are that they require long integration times, which can impact the potential latency benefits of SNNs. 


\noindent \textbf{Low-latency SNNs} (fast SNNs). SNN latency optimizations have received increasing scholarly attention in recent years~\cite{pfeiffer2018deep, deng2021optimal, bu2021optimal, li2022quantization, ding2021optimal}, and they can be coarsely divided to two categories: methods based on ANN-to-SNN conversion, and methods based on surrogate gradients. This research falls into the first category. Deng and Gu combine threshold balancing and soft-reset to correct conversion errors and show a significant reduction in the number of SNN simulation time steps needed~\cite{deng2021optimal}. The QCFS method~\cite{bu2021optimal} uses a quantized activation function to achieve ultra-low latency in SNNs. The QFFS method~\cite{li2022quantization}  proposes to reduce accuracy loss in low-latency SNNs by compressing information and suppressing noise.



\noindent \textbf{Dynamic Strategies}. Dynamic strategies~\cite{wu2018blockdrop, huang2017multi, chen2020dynamic, gao2018dynamic, almahairi2016dynamic, hua2019channel, lin2017runtime}, also known as dynamic networks, are network optimization algorithms that are input-dependent and can dynamically decide which part of the network to execute at runtime on a per-input basis.  The basic assumption in a dynamic strategy is that in real-world applications, examples are not equal for a given network model. Novel examples require the full model capacity to generate a reliable inference result, while simple examples can be solved confidently with fewer computing resources. Dynamic strategies optimize the inference of an ANN by taking both the model and input examples into considerations, and are promising for achieving fast and energy-efficient edge computing.

Compared to these studies on dynamic strategies, our method does not require any modifications to the internal structure of standard models or the addition of heavy auxiliary sub-networks as a dynamic controller to generate instructions. Instead, the proposed Dynamic Confidence can simply be calculated based on the neural network output during inference, bringing lower overhead and higher portability than other dynamic strategies. Also, most of the existing dynamic strategy research optimizes the number of channels and layers, while our method implicitly optimizes the activation precision, thanks to the event-based nature of SNNs. A visualization illustrating the dimensions of dynamic strategies can be found in the Supplementary Material.

\section{SNN preliminaries}
The detailed equations that describing spiking neuronal dynamics and ANN-to-SNN conversion are provided in the Supplementary Material. The following sections focus on introducing the main ideas of the proposed method in a comprehensive manner.

\section{Approach}\label{Approach}

The goal of our method is to allocate computing resources (specifically, inference latency) dynamically for each inference by using a confidence metric determined by model output. By this way, the time steps required for each inference are heterogeneous, and the averaged time steps per inference can potentially be reduced compared to using homogeneous time steps for all inferences.
The dynamic resource allocation mentioned above  can be modeled as a decision-making agent with two main points to consider:

\begin{enumerate}
  \item The decisions made regarding the allocation of computing resources to the model needs to facilitate fast and efficient inference while maintaining a model accuracy that is as high as possible. In other words, there is a minimal trade-off of accuracy for decreased latency.
  \item The decision-making mechanism itself needs to be lightweight and intelligible so as not to add unnecessary overhead and complexity to the model and to enable it to allow users to balance accuracy against latency.
\end{enumerate}

In our approach, we calculate the confidence associated with network output at runtime and use this confidence information to decide when to end inference in advance to get lower average latency per example without accuracy degradation. A confidence threshold is adopted to balance accuracy and latency. Section \ref{s3p1} describes the preliminary confidence concept, the quality of confidence, and confidence calibration. Section \ref{s3p2} explains the difference between confidence in SNNs and ANNs, highlights the challenges that must be addressed before utilizing confidence in SNNs, and presents our proposed solution. Section \ref{s3p3} presents how to construct a decision-making agent and exemplifies to use a Pareto Front to compute its threshold, which can be done with minimal cost. 

\subsection{Confidence}\label{s3p1}

In multi-class classification problems, for a given input example $X$, a label $Y$, and an artificial neural network $f(\cdot)$, the outputs are $f(X) = (\hat{Y}, \hat{P})$. $\hat{Y}$ is the predicted output label and $\hat{P} \in [0,1]$ is the confidence of this prediction.  Confidence $\hat{P}$ is an estimate of the ground truth inference correctness likelihood $P(\hat{Y} = Y)$, and it is usually calculated by Softmax in the multi-class classification problems discussed above. In some application scenarios where the safety and trustworthiness of the neural network models are central, such as medical diagnosis and autonomous driving, the confidence of the prediction is as crucial as its accuracy. The confidence can also be used to understand the interpretability and explainability of the neural network and can interact with other probabilistic models such as Bayesian neural networks.

\begin{figure}
\centering
\includegraphics[width=0.465\textwidth]{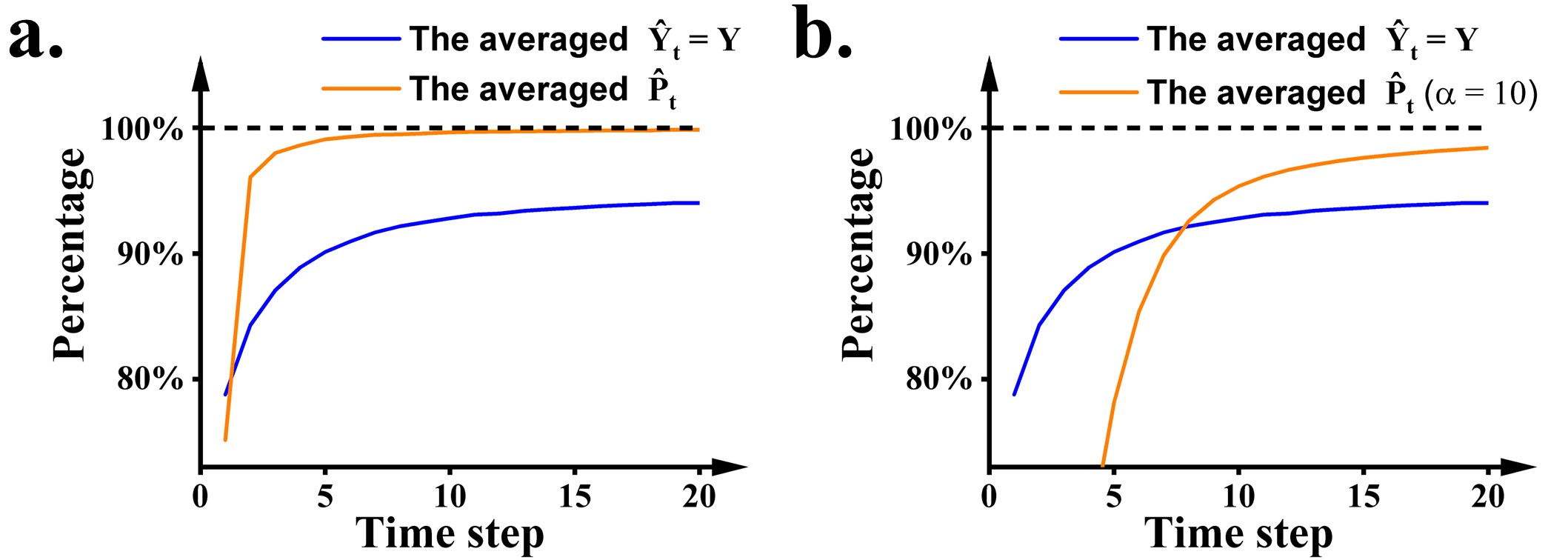} 
\caption{The trend of the averaged $\hat{Y_t} = Y$ and the averaged $\hat{P_t}$ over time $t$ on the test set of CIFAR-10. \textbf{a.} $\hat{P_t}$ saturates to 100\% quickly with time. \textbf{b.}. After introducing a scalar regulation parameter $\alpha$, $\hat{P_t}$ is softened.}
\label{fig2}
\end{figure}

Recognising the significance of $\hat{P}$, there is a growing body of literature that delivers calibration algorithms to improve the reliability of $\hat{P}$ ~\cite{ovadia2019can}. Note that calibration can only improve the quality of confidence; a perfectly calibrated model is impossible to achieve. Our approach applies temperature scaling~\cite{guo2017calibration}, a post-training calibration algorithm, to guarantee $\hat{P}$ to be of high quality and to keep the overhead of calibration low.

\subsection{Dynamic Confidence in Spiking Neural Networks}\label{s3p2}

After conducting ANN-to-SNN conversion, an ANN model $f(\cdot)$ will be converted into an SNN model $f_s(\cdot)$. Unlike a non-spiking neural network that only generates a single pair $(\hat{Y}, \hat{P})$, a spiking neural network $f_s(\cdot)$ with inputs $X_t$ will generate a series of outputs $(\hat{Y_t}, \hat{P_t})$ during inference, where $t \in \mathcal{T} =\{1, \ldots, T\}$ represents a time step during SNN inference. One feature of SNNs is that their prediction accuracy in the test set will increase with more time steps~\cite{pfeiffer2018deep}. This feature is associated with the event-based computing nature of SNNs: in each time step, only a fraction of the information is represented by spikes and processed by spiking neurons in an event-based manner; with more time steps, more information is accumulated and more reliable decision can be made. This well-known SNN feature means that: $\hat{Y}_{t_2} = Y$ is more likely to happen than $\hat{Y}_{t_1} = Y$ when $t_2 > t_1$. Here, we go one more step further and state that: If $\hat{P_t}$ is a good estimate of $\hat{Y_t} = Y$, $\hat{P}_{t_2}$ should also be more likely to be higher than $\hat{P}_{t_1}$ when $t_2 > t_1$.

To further clarify these, we plot the trend of $\hat{Y_t} = Y$ and $\hat{P_t}$ over time $t$ in Figure~\ref{fig2}.a., averaged on 10,000 test samples in CIFAR-10. The blue curve (the averaged $\hat{Y_t} = Y$) actually corresponds to the traditional accuracy-vs-latency response curve in SNNs \cite{pfeiffer2018deep}. The orange curve (the averaged $\hat{P_t}$) is an estimate of the correctness likelihood $P(\hat{Y_t} = Y)$, averaged on 10,000 test samples. We can see that both these curves show a monotonic increase with time $t$ with minor fluctuations, which is in line with the statements above. The same trend of these two curves suggests that $\hat{P_t}$ is a good estimator of $\hat{Y_t} = Y$: A low confidence value $\hat{P_t}$ indicates a low likelihood of correctness $\hat{Y_t} = Y$, and vice versa. Note that during SNNs inference, $\hat{Y_t} = Y$ is unknown because it needs to access the ground truth $Y$. However, this ground truth information can be implicitly fetched by estimating it by the confidence information $\hat{P_t}$. This provides a huge opportunity on optimizing SNNs by utilizing confidence. 

One obstacle must be overcome before optimizing SNNs with confidence. As shown in Figure~\ref{fig2}.a., the orange curve saturates to 100\% quickly with time $t$. In other words, an overconfident averaged $\hat{P_t}$ is produced after only a few time steps, after which there is little room for further improvement. We call this problem confidence saturation. Confidence saturation has a detrimental effect on our proposed method, as we want the $\hat{P_t}$ to act as a high-quality estimate of $\hat{Y_t} = Y$. A saturated $\hat{P_t}$ can no longer provide a good estimate of $\hat{Y_t} = Y$ since the saturated confidence is too trivial to distinguish from each other. The saturated confidence will also make the decision-making agent we are constructing later more sensitive to its parameter, which will be discussed in the following sections.

The confidence saturation problem is caused by the format of the SNN outputs. The outputs of a rate-coded SNN are accumulated spikes~\cite{li2022quantization} or, in some cases that need higher output precision, the membrane potential of integrate-but-not-fire neurons~\cite{li2021free, bu2021optimal}. In both situations, the output logits of SNNs $f_s(X_t)$ increase with $t$ continuously, which causes the saturated confidence $\hat{P_t}$ after applying Softmax to the SNN outputs. To prevent saturated confidence we introduce a scalar regulation parameter $\alpha$ to restrict the scale of the output logits in the SNNs, and calculate confidence according to

\begin{equation}
\hat{P_t} = \mathop {\max } ( \frac{e^{\mathbf{z}_{t}^{i}/\alpha}}{\sum_{j=1}^K e^{\mathbf{z}_{t}^{j}/\alpha}}) \ \ \ for\ i=1,2,\dots,K.
\end{equation}\label{equ1}

\noindent $\mathbf{z}_{t}$ is the output logits of the SNN, that is

\begin{equation}
\mathbf{z}_{t} =f_s(X_t)
\end{equation}

\noindent $\alpha$ serves to regulate the SNN's confidence level, ensuring it does not become overly saturated $\hat{P_t} \asymp 1$ before the end of the SNN simulation at time $T$. While various alpha values can fulfill this purpose, our selected alpha is intended to align the SNN's confidence value at $T$ with that of the ANN. This alignment guarantees a one-to-one match between the ANN and SNN models, allowing tools designed for ANN confidence to be applied to SNNs. With this consideration, $\alpha$ is set as $2^b-1$ when SNNs are built by QFFS \cite{li2022quantization}, where $b$ is the bit precision in all hidden layers. However, when SNNs are built by QCFS \cite{bu2021optimal}, deriving an exact alpha value becomes challenging 
because of the lack of existing research that can accurately estimate the effect of occasional noise \cite{li2022quantization} (also called unevenness error \cite{bu2021optimal}) on SNN output without accessing ground truth. The empirical solution we propose sets alpha to the same value as the number of simulation time steps $T$ when using QCFS. We plot the averaged $\hat{P_t}$ after applying scaling factor $\alpha$ in Figure~\ref{fig2}.b. It can be seen from this figure that after scaling the average $\hat{P_t}$ is softened.

\subsection{Optimizing Latency by Pareto Front}\label{s3p3}

\subsubsection{Optimization Targets}\label{s3p3p1}

Before conducting any optimization, we first clarify the targets of SNN latency optimization research. Certainly, the primary target of this type of research is achieving low latency in an SNN. On the other hand, we do not wish this low-latency SNN to have a low accuracy or to even be non-functional, which means that high accuracy should be considered as one of the optimization objectives as well. Hence, the goals of SNN latency optimization research should be at least twofold: achieving low latency and maintaining high accuracy. This is a typical multi-objective optimization problem, which allows a Pareto Front to be used to find efficient solutions.

\subsubsection{Proposed Method and Upper Bound}\label{s3p3p2}
In our proposed method, we dynamically optimize SNN latency based on the estimated correctness likelihood $\hat{P_t}$ for each example, while trying to ensure that $\hat{Y_t} = Y$. To accomplish this goal, we develop a simple decision-making agent that determines the optimal time step for terminating inference on a given sample. At runtime, a decision-making agent is placed at the output layer and it compares whether the estimated correctness likelihood $\hat{P_t}$ is higher than a confidence threshold $th_c$. If it is, a decision to terminate SNN inference will be made. For example, if $th_c = 0.6$ and $\hat{P_t} = [0.1, 0.3, 0.5, 0.7, 0.9]$ at the first 5 time steps for a given example, the SNN will be terminated in advance at the fourth time step as $\hat{P_4} = 0.7$, higher than $th_c = 0.6$. Different examples will have different termination times.

The optimal performance that can be achieved by our proposed dynamic strategy can be calculated in the following way: Firstly, assuming the ground truth label $Y$ is accessible during inference. In Dynamic Confidence, this corresponds to the assumption that our formulated confidence $\hat{P_t}$ is highly reliable and is a perfect estimate of $\hat{Y_t} = Y$. Secondly, the SNN inference will always be terminated for a given example at $t$ as soon as $\hat{Y_t} = Y$. This scenario represents the perfect capture of the correct termination time by the constructed decision-making agent. This upper-bound performance fully exploits dynamic strategies at runtime and represents the best possible performance that a given SNN can achieve.

We compared this upper-bound performance to the baseline performance on CIFAR-10 and ImageNet as displayed in Figure~\ref{fig1}. On CIFAR-10, the optimal solution achieved 95.1\% accuracy in 1.28~time steps and 96.7\% accuracy in 1.43~time steps, which is substantially better than the baseline that took 4~time steps to achieve 94.11\% accuracy. On ImageNet, the baseline performance was 72.68\% accuracy in 14~time steps, and it only surpassed 70\% accuracy in 4~time steps. In contrast, the optimal performance achieved 82.47\% accuracy in only 4.71~time steps. This significant performance gap suggests that there is potential for further improvements in SNNs by enabling runtime optimization. 


\begin{figure}
\centering
\includegraphics[width=0.465\textwidth]{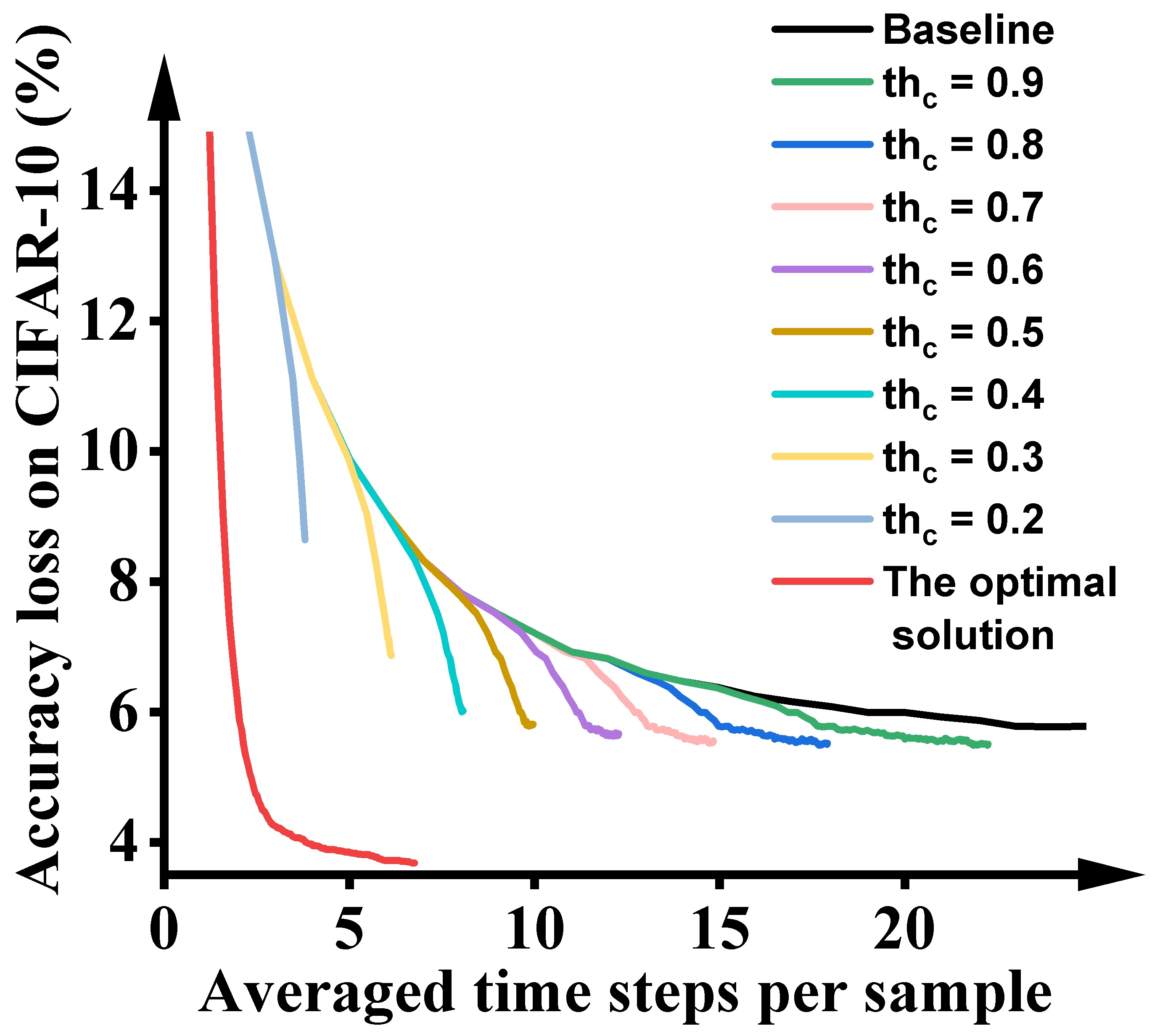} 
\caption{The Pareto Front of ResNet-18 on CIFAR-10 as well as the baseline performance (without a dynamic strategy) and the optimal solution (Assuming the ground truth is accessible).
Note that unlike the general form of Pareto Front, SNNs have a temporal dimension so their outputs are curves instead of points. The different settings of $th_c$ have different accuracy-vs-latency response curves.}
\label{fig4}
\end{figure}

\subsubsection{Calculating $th_c$ by Pareto Front}\label{Calculating by Pareto Front}

The value of the confidence threshold $th_c$ is crucial for the final inference performance. Intuitively, a high $th_c$ is a conservative dynamic strategy. It makes $\hat{P_t} > th_c$ happen less frequently and fewer examples will terminate early, limiting the latency improvements. In contrast, a low $th_c$ is a radical dynamic strategy, and making it easy for $\hat{P_t}$ to surpass $th_c$ and more examples will terminate early even though their output confidence $\hat{P_t}$ is low, bringing low latency but large accuracy drops. In other words, the selection of $th_c$ reflects the fundamental latency-accuracy trade-off in SNNs. To compute the precise confidence threshold $th_c$, we propose using a low-cost Pareto Front method. 

\begin{table*}
\caption{Latency advantages brought by Dynamic Confidence in 8 different experimental settings.}
\begin{center}
\begin{threeparttable}
\resizebox{\textwidth}{40mm}{

\begin{tabular}{c c c c c c c}

\toprule
\textbf{Dataset}& 
\textbf{Architecture}&
\textbf{Method}&\textbf{Acc(ANN)(\%)}&\textbf{Acc(SNN)(\%)}&\textbf{Averaged time steps}&\textbf{Latency saving}(\%)\\

\midrule
 \multirow{8}*{CIFAR-10} & \multirow{4}*{VGG-16}&QCFS &92.41 &92.50& 30 \\

&&\textbf{QCFS  + Dynamic Confidence}&92.41 &92.50& 12.69 &58\%\\
\cline{3-7}

&&QFFS  &92.41 &92.41& 6 \\

&&\textbf{QFFS  + Dynamic Confidence}& 92.41 &92.41& 3.17 &47\%\\
\cline{2-7}
&\multirow{4}*{ResNet-18}&QCFS &93.79 &94.27& 27 \\ 

&&\textbf{QCFS + Dynamic Confidence}&93.79 &\textbf{94.27}& 11.51&57\% \\

\cline{3-7}

&&QFFS &93.79 &94.11& 4 \\ 

&&\textbf{QFFS  + Dynamic Confidence}&93.79 &94.11& \textbf{2.52} &41\% \\ 

\cline{2-7}
&&Best Reported\cite{bu2021optimal}&95.52&93.96&4&\\
&&Best Reported\cite{li2022quantization}&93.12&93.14&4&\\

\midrule
\multirow{8}*{ImageNet}& \multirow{4}*{VGG-16}&QCFS& 72.40 & 73.30 & 74 \\

& &\textbf{QCFS + Dynamic Confidence}& 72.40 & \textbf{73.30} & 49.54 &33\% \\
\cline{3-7}
&&QFFS  &72.40 &72.52& 4 \\

&&\textbf{QFFS + Dynamic Confidence}  &72.40 &72.52& \textbf{2.86 } &29\% \\

\cline{2-7}

& \multirow{4}*{ResNet-50}&QCFS &72.60 &70.72& 128\\

&&\textbf{QCFS + Dynamic Confidence} &72.60 &70.72& 81.81&36\% \\
\cline{3-7}
&&QFFS &72.60 &73.17& 6 \\

&&\textbf{QFFS + Dynamic Confidence}  &72.60 &73.17& 4.42 &26\%\\
\cline{2-7}
&&Best Reported\cite{bu2021optimal}&74.29&72.85&64&\\
&&Best Reported\cite{li2022quantization}&71.88&72.10&4&\\
\hline

\end{tabular}}

 \begin{tablenotes}
        \footnotesize

    \item *A concurrent research \cite{hao2023bridging} based on QCFS reported a higher accuracy. However, note that Dynamic Confidence is a runtime latency optimization \\ method and can also apply to
            this SNN algorithm to further reduce its latency. 
      \end{tablenotes}
        \end{threeparttable}
\end{center}
\label{tab1}

\end{table*}

The optimization problem we are solving is essentially finding a series of $th_c$ values that achieve the highest accuracy for each latency. Computing a Pareto Front requires the search space to be finite, but there are infinitely many choices of $th_c$ in $[0, 1]$. To address this, we use a searching resolution of $0.1$ to discretize the search space of $th_c$, which is sufficient to get an appropriate $th_c$ that significantly reduces SNN inference time according to our experimental results. This suggests that our developed decision-making agent is highly robust to its parameter $th_c$. After discretizing the search space, there are only 11 settings of $th_c$. Then we draw the accuracy-vs-latency curve of $f_s(\cdot)$ with these 11 $th_c$ in Figure~\ref{fig4} (The setting of $th_c = 0$ is the same as $th_c = 0.1$ on CIFAR-10 and they are both terminated at the first time step with a very low accuracy of 78.76\%, and the setting of $th_c = 1$ is equivalent to the baseline that does not apply any dynamic strategy. Thus, we only show 8 candidate solutions in this figure.), by which a series of Pareto efficient choices of $th_c$ can be obtained. Then a $th_c$ which can give the lowest latency and no accuracy loss compared to the baseline solution is chosen from these 8 candidate solutions and used in Dynamic Confidence. For example, $th_c=0.6$ will be selected in Figure ~\ref{fig4}.

\section{Experiment}

\subsection{Experimental Setup}
\noindent \textbf{Datasets}. The proposed Dynamic Confidence is validated on CIFAR-10 and ImageNet. CIFAR-10 \cite{krizhevsky2009learning} contains 60,000 images, divided into 10 classes. There are 50,000 training images and 10,000 test images. The size of each image is 32x32 pixels and their format is RGB. ImageNet \cite{deng2009imagenet} is a large object recognition dataset, and the version we use in our experiments is ILSVRC-2012. ILSVRC-2012 has over one million labeled RGB examples, and 1000 object classes. It has 50,000 images for validation. 

\noindent \textbf{Network Architectures}.
For CIFAR-10 we experiment with VGG-16 and ResNet-18 and for ImageNet we experiment with VGG-16 and ResNet-50. Max-pooling is replaced by average-pooling to facilitate ANN-to-SNN conversion.

\noindent \textbf{QCFS and QFFS}. One of the essential features of the proposed Dynamic Confidence can be applied to a wide range of ANN-to-SNN conversion methods on the fly including the standard data-based normalization \cite{diehl2015fast} and 99.9\% data-based normalization \cite{rueckauer2017conversion}, resulting in significant reductions in inference latency. However, this paper is focused on exploiting the fast-response potential of SNNs, so we only highlight the results on ANN-to-SNN conversion methods that achieve ultra-low inference latency. Specifically, Dynamic Confidence is demonstrated on SNNs built using QCFS \cite{bu2021optimal} and QFFS \cite{li2022quantization}, both of which have demonstrated ultra-low latency with competitive accuracy on nontrivial datasets. For example, QCFS and QFFS have reported latencies of 64 and 4~time steps, respectively, achieving 72\% accuracy on ImageNet. Both methods reduce SNN latency by reducing the information in the ANN model (refer to \cite{li2022quantization} for a comprehensive introduction to why reducing ANN activation precision can lead to a reduction in SNN latency) and suppressing noise in the SNN model. QCFS achieves noise suppression by simulating longer to amortize the negative impact of noise, while QFFS generates negative spikes to correct noise. Some recent low-latency SNN algorithms \cite{hao2023reducing,hao2023bridging} also follow these two methods to amortize or correct noise.

In our experiments, we duplicate QCFS and QFFS by quantizing the activation in ANNs to 2-bit by LSQ \cite{esser2019learned} in all hidden layers and keeping the output layer at full precision. All weights are in full precision. We adopt the soft-reset IF neuron and its variant in SNNs as described in \cite{bu2021optimal, li2022quantization}, and the neurons in the output layer only accumulate input currents \cite{li2021free}. 
Detailed equations about LSQ, QCFS, and QFFS are provided in Supplementary Material.  

\noindent \textbf{Dynamic Confidence}. As described in the approach section, there are three steps to configure Dynamic Confidence:

\begin{enumerate}
    \item Calibration using an ANN.
    \item Scaling the output logits by $\alpha$ in the SNN converted from the calibrated ANN.
    \item Calculating the confidence threshold $th_c$ from the Pareto Front.
\end{enumerate}

Steps 1 and 3 are conducted on the same validation set. The validation set is collected randomly from the training set and its size is 5,000 on CIFAR-10 and 50,000 on ImageNet.

During inference, the confidence is calculated in SNN outputs and sent to the decision-making agent whose threshold is $th_c$. A binary decision of terminating the inference in advance is made if the confidence value surpasses $th_c$.

\noindent \textbf{Training Configurations}. The quantized ANNs are fine-tuned on the pre-trained full-precision ANN models for 60 epochs, with a momentum of 0.9 and a weight decay of $2.5 \times 10^{-5}$, and a loss function of cross entropy. The initial learning rate is 0.01 with an exponential learning rate decay. The batch size is 32. The training is implemented in PyTorch.

\subsection{Speed and Power Advantages of Using Dynamic Confidence}
While compromising accuracy can lead to even greater latency gains (by adopting a lower $th_c$, related results are shown in the Supplementary Material), our primary focus in the following sections is to optimize latency without sacrificing accuracy. Table~\ref{tab1} reports how much the average latency can be reduced by Dynamic Confidence in 8 different settings respectively. Note that after applying Dynamic Confidence to SNNs, the inference latency is heterogeneous for different examples, so the averaged latency is not necessarily an integer. The results show that Dynamic Confidence can bring a latency reduction of 41\% to 58\% on CIFAR-10, and a latency reduction of 26\% to 36\% on a more challenging dataset ImageNet without sacrificing accuracy. These substantial improvements in latency suggest the obvious benefits of adopting Dynamic Confidence to allow heterogeneous terminating time. Moreover, note that both QCFS and QFFS already provide ultra-fast solutions on rate-coded SNNs. Even though, applying Dynamic Confidence on top of these two fast SNN solutions can still render significant latency reduction.

The power cost of an SNN is roughly kept stable for each time step for a given SNN run on GPUs. Given this observation, reducing the averaged time steps required by an SNN at runtime can also bring nontrivial gains in power efficiency. For example, by reducing the averaged latency from 30 time steps to 12.69 time steps in CIFAR-10; the power consumption of SNN simulations can be roughly reduced by 58\%. The overhead of applying Dynamic Confidence may dilute this gain in power a little bit, which is discussed in Section \ref{Speedup and Overhead}.



















\subsection{Comparison with State-of-the-Art Methods}

State-of-the-art performance of fast SNNs on CIFAR-10 and ImageNet is listed in Table~\ref{tab1}. From this table, we can see that on CIFAR-10, the performance achieved by applying Dynamic Confidence in QFFS (94.11\% in 2.52 time steps)
outperforms the state-of-the-art results both in accuracy and latency. On ImageNet, we report an accuracy of 72.52\% in 2.86~time steps, outperforming state-of-the-art results as well.

Not all neuromorphic hardware and SNN algorithms support the negative spikes used in QFFS. Thus, we also benchmark Dynamic Confidence on QCFS which does not use negative spikes. As shown in this table, even without negative spikes, Dynamic Confidence can still achieve an accuracy of 73.30\% on ImageNet with an averaged latency of 49.54~time steps, outperforming the state-of-the-art method (72.85\% in 64 time steps) by a significant margin.

\subsection{Latency Distributions After Enabling Heterogeneous Terminating Time}
\begin{figure}
\centering
\includegraphics[width=0.465\textwidth]{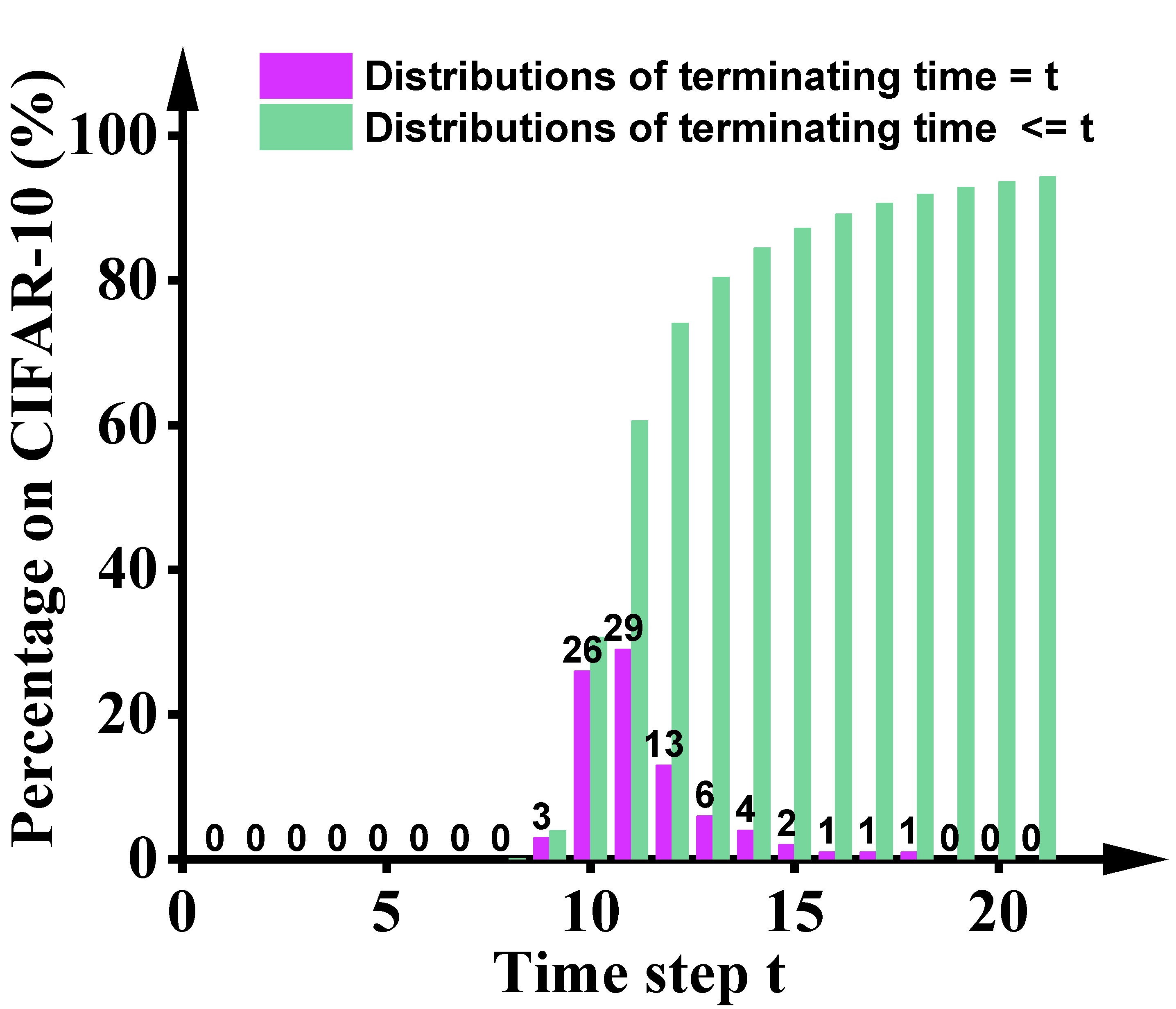} 
\caption{The distributions of terminating time after applying Dynamic Confidence on CIFAR-10. The network architecture is ResNet-18. We can see that instead of a fixing inference latency for each sample, different samples can have heterogeneous terminating times when adopting Dynamic Confidence.}
\label{fig5}
\end{figure}

We record the distributions of terminating time in Figure~\ref{fig5}, to visualize how many samples are terminated by Dynamic Confidence early and facilitate a better understanding of Dynamic Confidence. The purple histogram shows the percentage of samples terminated at a time point $t$ and the green histogram is the total terminated samples before $t$ and at $t$. The dataset is CIFAR-10 and the network architecture is ResNet-18. It is apparent that the terminating time varies for different samples. There is a significant percentage of samples terminated at the time point $10$ and $11$ (26\% and 29\%), and up to 94.97\% samples are terminated before the time point $22$. Even though Dynamic Confidence terminates such a large percentage of samples at very early time steps, its inference accuracy is still guaranteed to be as high as 94.27\%. This suggests that the terminating decision made by Dynamic Confidence is highly reliable.

\subsection{Overhead of Dynamic Confidence}\label{Speedup and Overhead}

Dynamic Confidence has been demonstrated to provide significant improvements in terms of latency, as discussed in previous sections. Also, Dynamic Confidence is on-the-fly so it is a promising tool for further optimizing latency and spike counts on other low-latency SNN algorithms as well. However, to ensure that its benefits are maximized, it is important to be mindful of the potential overhead introduced by implementing Dynamic Confidence, as excessive costs can offset the benefits it provides. The overhead of Dynamic Confidence at the configuration phase is trivial, which is illustrated in the Supplementary Material. This section focuses on analyzing the overhead of Dynamic Confidence at runtime, particularly in the context of real-world applications where it holds greater significance.

At runtime, the main computational overhead of Dynamic Confidence is confidence calculation by Softmax, whose computational complexity is $\mathcal{O}(TL)$. $T$ is the time step during SNN simulation and $L$ is the number of output neurons. Benefiting from the rapid developments of low-latency SNN algorithms, $T$ has been reduced significantly in recent years. For example, when conducting Dynamic Confidence with QFFS and CIFAR-10, $T$ can be as small as $4$. As for $L$, since the main application scenario of SNNs is small-scale edge-computing tasks, $L$ is limited in most cases. For instance, $L$ is $10$ in CIFAR-10 and $11$ in DVS-Gesture. Therefore, compared to the 26\% to 58\% gains in latency reported in previous sections, the runtime overhead of Dynamic Confidence is insignificant, especially when applying Dynamic Confidence on low-latency SNNs and edge AI applications.
While low overhead is crucial to our discussion, it can be justifiable to tolerate higher overhead if a dynamic strategy can achieve performance that approaches the upper bound of an SNN, such as the 82.47\% accuracy on ImageNet in 4.71 time steps depicted in Figure~\ref{fig1}.


\subsection{Actual Inference Latency Reduction Recorded on GPUs}

Table~\ref{tab2} presents the average latency per inference for SNN simulations on GPUs. The simulations were conducted on a NVIDIA-V100 GPU, with a batch size of 1. It is important to note that the reported latency in this table is a measurement of the actual inference speed of SNN models on GPUs, whereas the latency gains shown in Table~\ref{tab1} are calculated by recording the average simulated time step required per inference.

 \begin{table}[h]
\caption{Latency per inference with/without Dynamic Confidence. The unit is seconds. The lower the better.}
\begin{center}
\begin{tabular}{ccc}
\hline
Setting& no DC & DC 
\\
\hline
CIFAR-10, VGG-16, QCFS&0.097& 0.047
\\

CIFAR-10, VGG-16, QFFS&0.036&0.017
\\
CIFAR-10, ResNet-18, QCFS&0.127& 0.047
\\

CIFAR-10, ResNet-18, QFFS&0.027&0.022 
\\

ImageNet, VGG-16, QCFS&0.315& 0.213
\\

ImageNet, VGG-16, QFFS&0.030& 0.021
\\

ImageNet, ResNet-50, QCFS&1.708& 0.932
\\

ImageNet, ResNet-50, QFFS&0.116& 0.104
\\

\hline

\end{tabular}

\label{tab2}
\end{center}
\end{table}

\subsection{Spike Count Reduction}\label{gpu}

Similar to other SNN algorithm research, our proposed method is demonstrated and evaluated through the simulations on GPUs, the real power saving when running on neuromorphic hardware is unknown to us. Given that neuromorphic hardware's energy consumption is dominated by the multicast routing of spikes, we report the data on the spike count reductions achieved by our method. This is to provide insight into its potential power-saving benefits. 

Table~\ref{tab3} reveals that the implementation of Dynamic Confidence significantly lowers spike counts, most notably for QCFS. In contrast, when using QFFS configurations, the reduction is more modest. This indicates that although Dynamic Confidence invariably lowers spike counts, its effectiveness can differ based on the chosen architecture and specific settings.

 \begin{table}[h]
\caption{Spike counts per neuron per inference with/without Dynamic Confidence. The lower the better.}
\begin{center}
\begin{tabular}{ccc}
\hline
Setting& no DC & DC 
\\
\hline
CIFAR-10, VGG-16, QCFS&3.21& 1.35 
\\

CIFAR-10, VGG-16, QFFS&0.32&0.31
\\
CIFAR-10, ResNet-18, QCFS&3.02& 1.29 
\\

CIFAR-10, ResNet-18, QFFS&0.34&0.25 
\\

ImageNet, VGG-16, QCFS&12.94& 8.66
\\

ImageNet, VGG-16, QFFS&0.52& 0.45
\\
ImageNet, ResNet-50, QCFS&21.63& 13.84
\\

ImageNet, ResNet-50, QFFS&0.53& 0.51
\\

\hline

\end{tabular}

\label{tab3}
\end{center}
\end{table}

\section{Towards Greater Opportunities}

\noindent \textbf{Visualizing and Approaching the ``Potential" of SNNs}. Section \ref{s3p3p2} presents a method to calculate the upper-bound performance of an SNN, which opens up numerous exciting possibilities. With this approach, we can visualize the potential of an SNN when its correct terminating time is accurately captured and compare it with that of another SNN. Figure~\ref{fig1} depicts that SNNs can achieve highly competitive accuracy (in some cases, even surpass the capacity of the original ANN model) in just a few time steps. This potential can be explained by viewing the SNN as an ensemble network. The input-output response curve of spiking neurons in an SNN is highly dependent on its terminating time point. By stopping the simulation at different time steps, we effectively select a different subnetwork with a distinct input-output response curve. For example, if an SNN simulates 10 time steps, it is essentially an ensemble of 10 subnetworks. These ten subnetworks share the same weights, but their input-output response curve is different. For different input samples, we can select the most appropriate subnetwork from these 10 subnetworks to do inference, simply by terminating this SNN at the corresponding time point. Therefore, the temporal dimension makes SNNs a highly efficient ensemble method. 

SNNs have huge latent potential, the key question is how to unleash this. In this paper, we exemplify a method to partly unlock this potential by 1). formulating a high-quality estimation of the ground truth and 2). constructing a highly-reliable decision-making agent. 


\noindent \textbf{Further Applications of Dynamic Confidence}. To improve the implemented performance of Dynamic Confidence on neuromorphic hardware, it may be desirable to use an alternative mechanism to the Softmax function. One option is to use a $k$-winners-take-all (k-WTA) spiking neural architecture whereby lateral inhibition between output neurons gates outputs. This could make Dynamic Confidence more amenable to implementation directly on neuromorphic hardware. Dynamic Confidence is possible to be applied to SNNs with timing-based encoding methods such as rank order coding \cite{garcia2007evolution}, latency coding \cite{pan2019neural}, and Time-to-First-Spike Coding \cite{park2020t2fsnn}, and to SNNs with surrogate gradients training methods to further leverage the temporal information in SNNs. In timing-based encoding methods, they may need a new metric than Softmax to represent confidence. It's worth noting that, even when using a superior confidence metric, the upper-bound performance achieved with this metric still aligns with our calculations.


\section{Conclusion}

This study critically examines the ways of applying dynamic strategies to spike-based models to optimize inference latency by Dynamic Confidence. Essentially, we formulate confidence in SNNs and use it to decide whether terminate inference or wait more time steps to accumulate more evidence and get a more reliable prediction. The major challenge of introducing the concept of confidence to SNNs is that, unlike confidence in ANNs, confidence in SNNs evolves with time, so some modifications need to be applied to leverage the information carried by confidence. A simple decision-making agent is then constructed to decide when to stop inference. Dynamic Confidence improves the performance of SNNs in the aspect of latency and power for exploring their potential on becoming a strong candidate for low-power low-latency edge computing.

\section{Acknowledgments
}
The authors would like to acknowledge numerous
helpful comments from the reviewers. Chen Li thanks Luziwei Leng and Zhaofei Yu for helpful feedback on the first draft of this paper. We also recognize the support from Research IT and the utilization of the Computational Shared Facility at The University of Manchester.

{\small
\bibliographystyle{ieee_fullname}
\bibliography{main}
}

\renewcommand{\thesection}{S\arabic{section}}
\setcounter{section}{0}
\beginsupplement
\section{Additional Figures with Other Settings}

\begin{figure}
\centering
\includegraphics[width=0.47\textwidth]{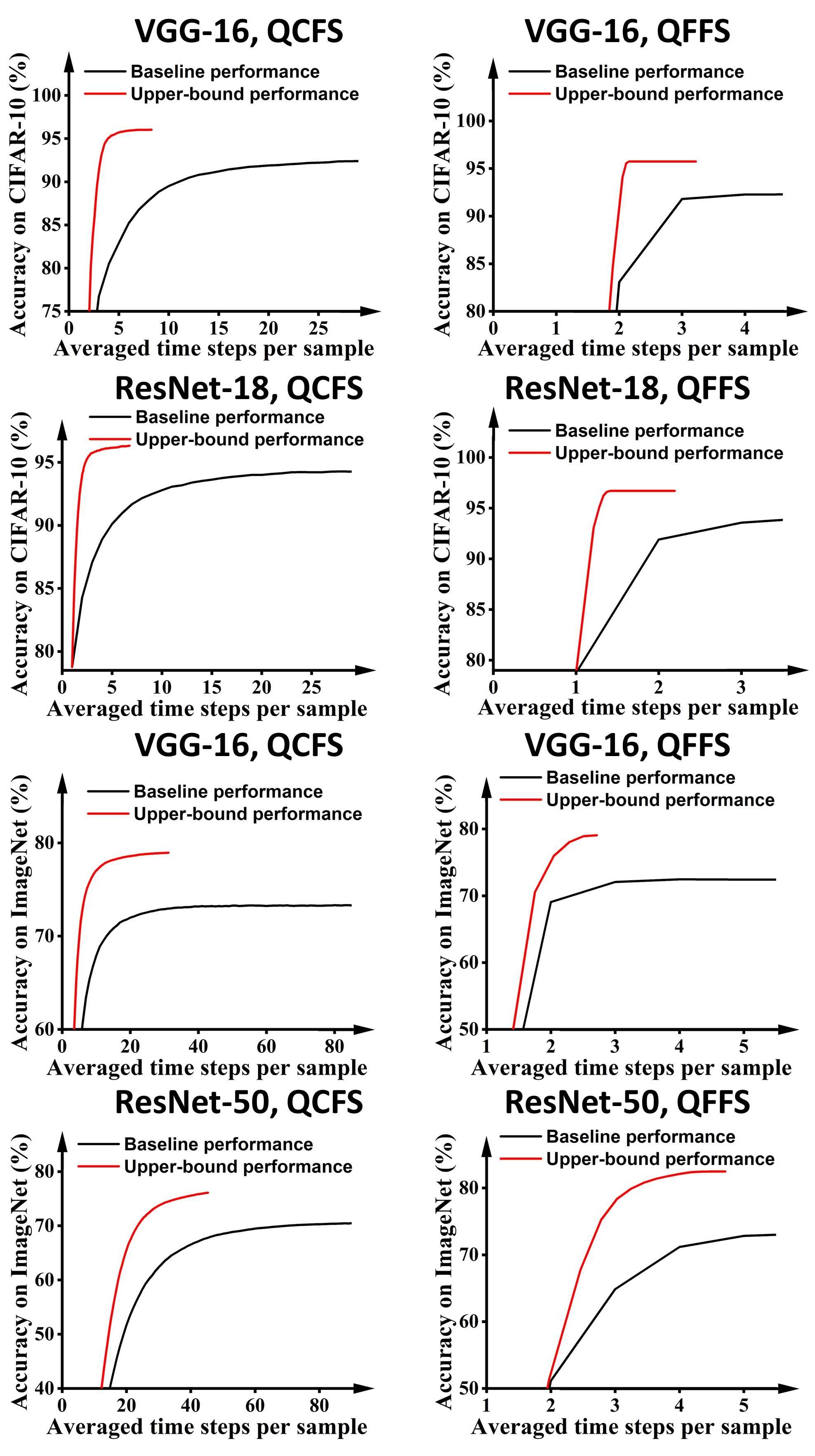} 
\caption{The upper-bound performance of SNNs when fully utilizing dynamic strategies at runtime, as shown by the red curves. The black curves represent baseline SNN performance without dynamic strategies.}
\label{sf2}
\end{figure}

Figure~\ref{sf2} illustrates the upper-bound performance of SNNs across all eight settings.

\section{ Visualization of Dynamic Strategies}

\begin{figure}
\centering
\includegraphics[width=0.47\textwidth]{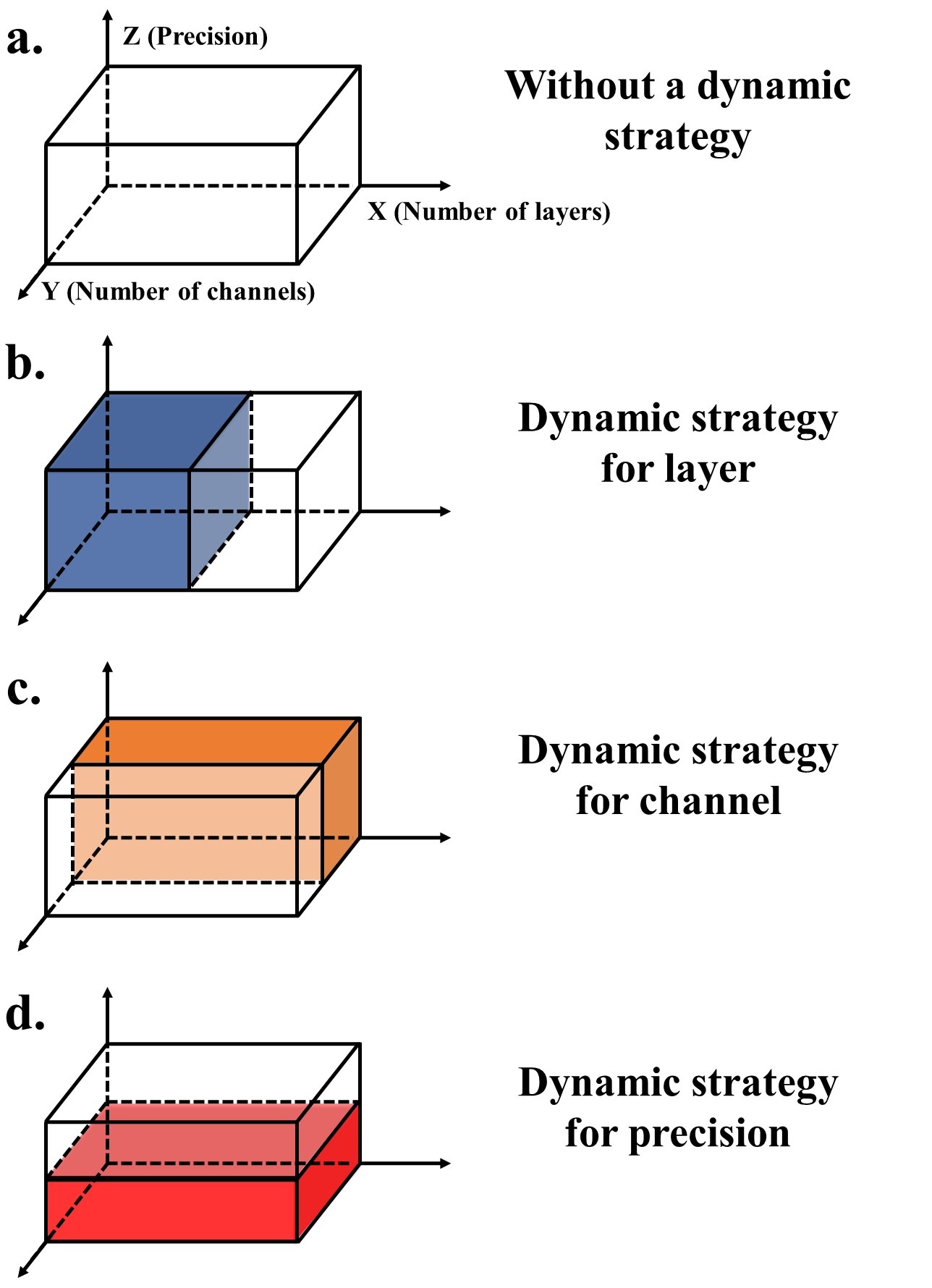} 
\caption{The visualization of dynamic strategies in a three-dimensional space. The cuboid in \textbf{a.} represents a neural network model without applying any dynamic strategies. For each example, the whole model capability will be used to generate outputs. \textbf{b. c. d.} apply dynamic strategies in the dimension of layer, channel, and precision, respectively. These dynamic strategies can adapt to different examples. Instead of the full model capacity, only a portion of computational resources,
just enough to generate reliable outputs (The colored part of the cuboid), will be allocated for a given example. Each example has heterogeneous resource allocation.}
\label{sf1}
\end{figure}

Dynamic strategies can be visualized in a three-dimensional space with three axes representing the number of layers, the number of channels, and precision, see Figure~\ref{sf1}. The majority of research into dynamic strategies, such as BlockDrop~\cite{wu2018blockdrop}, feature boosting and suppression~\cite{gao2018dynamic}, and Dynamic Convolution~\cite{chen2020dynamic}, dynamically skip either layers (see Figure~\ref{sf1}.b.) or channels (see Figure~\ref{sf1}.c.), optimizing dimensions one and two respectively. 

Some work on dynamic strategies (see Figure~\ref{sf1}.d.) in dimension three, the optimization of precision at runtime, has been carried out. Song et al. proposed dynamic region-based quantization (DRQ) in which a predictor classified regions in an input image as requiring 8~bit or 4~bit weights~\cite{songDRQDynamicRegionbased2020}. Huang et al. present a similar method in which the less significant bits of an activation representation can be dynamically masked to allow on-the-fly changes to the multiplications carried out~\cite{huangStructuredDynamicPrecision2022}. So far dynamic strategies for precision are based on selecting a precision for a certain value from a discrete set of integer precisions. This is in contrast with the work presented in this paper where the effective activation precision can be varied smoothly based on the inference confidence by virtue of using an SNN model.

\section{Spiking Neurons and ANN-to-SNN Conversion}

The neuronal dynamics of integrate-and-fire neuron in SNNs is controlled by these two equations below:

\begin{equation}\label{seq1}
\bm{u}_{t}^{l} = \bm{u}_{t-1}^{l}(1-\bm{z}_{t-1}^{l})+\widetilde{\bm{W}}^l\bm{z}_{t}^{l-1}+\widetilde{\bm{B}}^l 
\end{equation}

\begin{equation}\label{seq2}
\bm{z}_{t}^{l} =\Theta(\bm{u}_{t}^{l}-\bm{th}^{l})
\end{equation}

\noindent In these equations, $\bm{u}_{t}^{l}$ is the membrane potential of spiking neurons in layer $l$ at time $t$. $\bm{u}_{t-1}^{l}$ is the membrane potential at the previous time step $t-1$.
$\widetilde{\bm{W}}^l$ is the weight and $\widetilde{\bm{B}}^l$ is the bias in layer $l$. $\Theta(\cdot)$ denotes the Heaviside step function, and it returns $1$ if the value in the bracket is higher than zero; otherwise it returns $0$. $\bm{th}^{l}$ is the threshold in layer $l$, and $\bm{z}_{t}^{l}$ is the output spike in this layer at time $t$.

ANN-to-SNN conversion is using spiking neurons to achieve the computations that happen in artificial neurons. The information processing in the artificial neurons in layer $l$ is 

\begin{equation}\label{seq3}
\bm{y}^{l} = \sigma( \bm{W}^{l} \bm{y}^{l-1}+\bm{B}^l)
\end{equation}

\noindent where $\sigma(\cdot)$ is the ReLU activation function, $\bm{W}^l$ and $\bm{B}^l$ denote the weight and the bias in layer $l$, $\bm{y}^{l}$ is the output of layer $l$, and $\bm{y}^{l-1}$ is the output in the previous layer.

The original method to conduct ANN-to-SNN conversion is called data-based normalization \cite{diehl2015fast}, where the SNN parameters are calculated by

\begin{equation}
\widetilde{\bm{W}}^l = \frac{\lambda^{l-1}\bm{W}^l}{\lambda^l} \label{equ10}
\end{equation}

\begin{equation}
\widetilde{\bm{B}}^l = \frac{\bm{B}^l}{\lambda^l} \label{equ11}
\end{equation}

\begin{equation}
\bm{th}^{l} = 1 \label{equ12}
\end{equation}

\noindent In these equations,  $\lambda^l$ and $\lambda^{l-1}$ are the maximum ANN activation value in layer $l$ and the previous layer $l-1$ respectively.

\begin{table*}
\caption{Latency advantages brought by Dynamic Confidence in 8 different experimental settings. The accuracy is compromised for better latency.}
\begin{center}
\begin{threeparttable}
\resizebox{\textwidth}{35mm}{

\begin{tabular}{c c c c c c c}

\toprule
\textbf{Dataset}& 
\textbf{Architecture}&
\textbf{Method}&\textbf{Acc(ANN)(\%)}&\textbf{Acc(SNN)(\%)}&\textbf{Averaged time steps}&\textbf{Latency saving}(\%)\\

\midrule
 \multirow{8}*{CIFAR-10} & \multirow{4}*{VGG-16}&QCFS &92.41 &92.31& 27 \\

&&\textbf{QCFS  + Dynamic Confidence}&92.41 &92.31& 10.85 &60\%\\
\cline{3-7}

&&QFFS  &92.41 &92.31& 6 \\

&&\textbf{QFFS  + Dynamic Confidence}& 92.41 &92.31& 2.87 &52\%\\
\cline{2-7}
&\multirow{4}*{ResNet-18}&QCFS &93.79 &94.00& 19 \\ 

&&\textbf{QCFS + Dynamic Confidence}&93.79 &\textbf{94.00}& 8.03&63\% \\

\cline{3-7}

&&QFFS &93.79 &93.79& 4 \\ 

&&\textbf{QFFS  + Dynamic Confidence}&93.79 &93.79& \textbf{2.16} &46\% \\ 


\midrule
\multirow{8}*{ImageNet}& \multirow{4}*{VGG-16}&QCFS& 72.40 & 73.00 & 33 \\

& &\textbf{QCFS + Dynamic Confidence}& 72.40 & \textbf{73.00} & 27.90 &15\% \\
\cline{3-7}
&&QFFS  &72.40 &72.52& 4 \\

&&\textbf{QFFS + Dynamic Confidence}  &72.40 &72.40& \textbf{2.78 } &31\% \\

\cline{2-7}

& \multirow{4}*{ResNet-50}&QCFS &72.60 &70.50& 94\\

&&\textbf{QCFS + Dynamic Confidence} &72.60 &70.50& 67.22&28\% \\
\cline{3-7}
&&QFFS &72.60 &72.56& 6 \\

&&\textbf{QFFS + Dynamic Confidence}  &72.60 &72.56& 3.79 &37\%\\
\cline{2-7}

\hline

\end{tabular}}

        
        \end{threeparttable}
\end{center}
\label{stab1}

\end{table*}

\section{LSQ}
LSQ, or Learned Step Size Quantization \cite{esser2019learned}, is a method to build low-precision ANNs. Compared with other quantization-aware training approaches, the quantization scale factors in LSQ are trainable, which benefits to better mapping from full-precision values to quantized values.

A quantization operator is defined by

\begin{equation}\label{seq7}
	\hat{v} = s\lfloor clip( \frac{v}{s}, -Q_N, Q_P) \rceil,
\end{equation}

\noindent where $v$ is the full-precision value before quantization, $s$ is a quantization scale factor. $clip(z, -Q_N, Q_P)$ clips $z$ to the range of $(-Q_N, Q_P)$. For example, in 2-bit activation quantization, this range is $(0, 3)$. $\lfloor z \rceil$ is rounding operation that rounds $z$ to the nearest integer. 

During training, the gradient of $\frac{\partial{\hat{v}}}{\partial{v}}$ is defined by applying straight-through estimators~\cite{neftci2017event}, which is

\begin{equation}\label{seq8}
\frac{\partial{\hat{v}}}{\partial{v}} =
\begin{cases}
1			& \text{if $-Q_N < \frac{v}{s} < Q_P$} \\
0								& \text{if $\frac{v}{s} \le -Q_N$} \\
0.							& \text{if $\frac{v}{s} \ge Q_P$} \\
\end{cases}
\end{equation}

\noindent The gradient of $\frac{\partial{\hat{v}}}{\partial{s}}$ is

\begin{equation}\label{seq9}
\frac{\partial{\hat{v}}}{\partial{s}} =
\begin{cases}
-\frac{v}{s} + \lfloor \frac{v}{s} \rceil			& \text{if $-Q_N < \frac{v}{s} < Q_P$} \\
-Q_N									& \text{if $\frac{v}{s} \le -Q_N$} \\
Q_P,									& \text{if $\frac{v}{s} \ge Q_P$} \\
\end{cases}
\end{equation}

\noindent which is also calculated by applying straight-through estimators. The detailed calculations are shown below.

\begin{equation}\label{seq10}
    \begin{aligned}
	\frac{\partial{\hat{v}}}{\partial{s}} &= s\lfloor \frac{v}{s} \rceil 
     & \text{if $-Q_N < \frac{v}{s} < Q_P$} \\
 	 &= s^{'}\lfloor \frac{v}{s} \rceil + s\lfloor \frac{v}{s} \rceil^{'} \\
    	 &= \lfloor \frac{v}{s} \rceil + s(\lfloor \cdot  \rceil^{'} (\frac{v}{s})^{'}) \\
        &= \lfloor \frac{v}{s} \rceil + s (\frac{v}{s})^{'} \\
        &= \lfloor \frac{v}{s} \rceil + s (-\frac{v}{s^2})^{'} \\
        &= \lfloor \frac{v}{s} \rceil -\frac{v}{s} \\
    \end{aligned}
\end{equation}

\section{QCFS and QFFS}
In this Section, we first formulate the target ANN computations to be simulated in SNNs by QCFS and QFFS.  After replacing ReLU in Equation ~\ref{seq3} with the activation quantization operator defined in Equation ~\ref{seq7}, the computation in a layer of ANN becomes

\begin{equation}\label{seq11}
	\bm{y}^{l} = s^l\lfloor clip( \frac{\bm{W}^{l} \bm{y}^{l-1}+\bm{B}^l}{s^l}, -Q_N, Q_P) \rceil.
\end{equation}

\noindent Our experiments use 2-bit activation quantization where $Q_N$ is $0$ and $Q_P$ is $3$, so this equation becomes 

\begin{equation}\label{seq12}
	\bm{y}^{l} = ^l\lfloor clip( \frac{\bm{W}^{l} \bm{y}^{l-1}+\bm{B}^l}{s^l}, 0, 3) \rceil.
\end{equation}

\textbf{QCFS}. The spiking neuronal model used in QCFS is the soft-reset integrate-and-fire neuron, which is

\begin{equation}\label{seq13}
\bm{u}_{t}^{l} = \bm{u}_{t-1}^{l}+\widetilde{\bm{W}}^l\bm{z}_{t}^{l-1}\bm{th}^{l-1}+\widetilde{\bm{B}}^l-\bm{z}_{t-1}^{l}\bm{th}^{l}, 
\end{equation}

\begin{equation}\label{seq14}
\bm{z}_{t}^{l} =\Theta(\bm{u}_{t}^{l}-\bm{th}^{l}).
\end{equation}

\noindent There are three parameters ($\widetilde{\bm{W}}$, $\widetilde{\bm{B}}^l$, and $\bm{th}^{l}$) that need to be calculated according to ANN parameters. The equations are provided below:

\begin{equation}\label{seq15}
\widetilde{\bm{W}}^1 = \bm{W}^l
\end{equation}

\begin{equation}\label{seq16}
\widetilde{\bm{B}}^1 = \bm{B}^l 
\end{equation}

\begin{equation}\label{seq17}
\bm{th}^l =   3{s^l}.
\end{equation}

\noindent Note that, unlike QFFS, QCFS does not have a maximum spike count limitations per neuron ($Z\_max$ in Equation \ref{seq19}) to model the clipping operation on $Q_P$ in the target ANN (Equation \ref{seq12}). As a result, after the simulation of QFFS stops, QCFS can continue to simulate and at a time point, its accuracy will surpass that of QFFS. 

\textbf{QFFS}. The spiking neuronal model used in QFFS is the soft-reset integrate-and-fire neuron with negative spikes, which is defined by the equations below:

\begin{equation}\label{seq18}
\bm{u}_{t}^{l} = \bm{u}_{t-1}^{l}+\widetilde{\bm{W}}^l\bm{z}_{t}^{l-1}\bm{th}^{l-1}+\widetilde{\bm{B}}^l-\bm{z}_{t-1}^{l}\bm{th}^{l} 
\end{equation}

\begin{equation}\label{seq19}
\bm{z}_{t}^{l} =\Theta(\bm{u}_{t}^{l}-\bm{th}^{l})\Theta(Z_{max}-\bm{Z}_{t-1}^{l})-\Theta(-\bm{u}_{t}^{l})\Theta(\bm{Z}_{t-1}^{l}) 
\end{equation}

\begin{equation}\label{seq20}
\bm{Z}_{t}^{l} =\bm{Z}_{t-1}^{l}+\bm{z}_{t}^{l}. 
\end{equation}

\noindent $Z_{max}$ is the maximum spike count limitation, and $\bm{Z}_{t}^{l}$ records the accumulated spike counts. The ANN-to-SNN conversion is achieved according to the equations below:

\begin{equation}
\widetilde{\bm{W}}^l = \bm{W}^l \label{seq21}
\end{equation}

\begin{equation}
\widetilde{\bm{B}}^l = \bm{B}^l \label{seq22}
\end{equation}

\begin{equation}
\bm{th}^{l} = 3s^l \label{seq23}
\end{equation}

\begin{equation}
Z_{max} = 3\label{seq24}
\end{equation}

\section{Configuration Overhead of Dynamic Confidence}

The configurations of Dynamic Confidence have three steps as illustrated in main sections. In step 1 an ANN model is run for one epoch on a small valid set for calibration, this cost is about 600 times lower than the whole training cost and it can be faithfully ignored. Step 2 calculates a scale factor $\alpha$ to use in inference, which does not bring any overhead. In step 3 the confidence threshold $th_c$ is calculated by Pareto Front. Generally, Pareto Front can be very expensive since Pareto Front replies on high-resolution search space to ensure the performance of the Pareto-optimal one. However, Dynamic Confidence is very power efficient as our proposed method is highly robust to the search resolution of $th_c$
when applying Pareto Front. This robustness may come from step 2 which softens the confidence value before computing Pareto Front. Second, we emphasize that even applying a high search resolution of $th_c$ such as 0.0001, the computational complexity of the Pareto Front is still trivial, only if exploiting the monotonicity of thresholded classifications, similar to the trick used when calculating ROC curve in machine learning\cite{fawcett2006introduction}. Specifically, any input samples that are terminated by Dynamic Confidence with respect to a given $th_c$ will be terminated for all lower thresholds. Leveraging this property, a high-search-resolution Pareto Front can be created by running an SNN once on a valid set, and then, on the collected SNN outputs conducting a linear scan of $th_c$ (which can be done in a few seconds on a laptop). In summary, the main overhead of Dynamic Confidence at the configuration phase is running an ANN and an SNN on a valid set, which is trivial and can be ignored. 

\section{More Results with Compromised Accuracy}

Table~\ref{stab1} displays the results of compromising the accuracy of SNNs to achieve better latency.

\section{More Related Work}



\noindent \textbf{Model Compression}. In order to deploy neural network algorithms on resource-constrained edge devices and to meet the edge restraints on model size, inference latency, and power cost, several model compression techniques have been proposed such as quantization~\cite{han2015deep, courbariaux2015binaryconnect, lin2022device, esser2019learned}; structured and unstructured pruning~\cite{he2017channel, han2015deep}; compact network architecture design~\cite{howard2017mobilenets, huang2017densely, iandola2016squeezenet}; neural architecture search (NAS)~\cite{elsken2019neural, liu2018progressive}; and algorithm-hardware co-design~\cite{han2017efficient}.

The model compression techniques introduced above optimize the neural network model in an offline manner and will not adapt to inputs during execution. This paper, on the other hand, is focused on runtime dynamic optimization methods (also called dynamic strategies or adaptive computation) that can choose different computing strategies for different input images. Our method is orthogonal to these model compression techniques and could be combined with them to further improve SNN performance.

\section{More Discussions}

Except for Pareto Front, $th_c$ can be calculated by such as applying reinforcement learning or SGD learning (SGD learning needs to define the exact form of the gradients of $th_c$, where surrogate gradients may be desirable). These methods may be more effective to find better thresholds, but this paper 
emphasizes the efficiency of our method, so does not involve learning and keep it post-hoc.

Rate-coding is generally considered inefficient in its use of spikes~\cite{davidson2021comparison}. However, the method we have developed exploits the smoothly variable precision available with rate-coding to process incoming data incrementally, opening the door to saving compute resources by early exiting from the inference pipeline.
The efficiency of this kind of approach in terms of the number of operations is dependent on the number of input examples that are `easy' and allow for early exiting. In other words, Applying Dynamic Confidence on less challenging datasets such as MNIST and CIFAR-10 can render greater improvements than on ImageNet.

In applications where model input data are streams of events directly from sensors, such as with event camera input or event-based scintillation detectors, Dynamic Confidence can provide a reduction in the latency between event generation and model output. This makes Dynamic Confidence a particularly attractive candidate algorithm for exploration in low-power low-latency edge computing.

\end{document}